\pdfoutput=1
\documentclass[11pt]{article}

\usepackage[final]{acl}

\usepackage{times}
\usepackage{latexsym}

\usepackage[T1]{fontenc}
\usepackage[utf8]{inputenc}

\usepackage{microtype}

\usepackage{inconsolata}

\usepackage{graphicx}

\title{Towards an Analysis of Discourse and Interactional Pragmatic Reasoning Capabilities of Large Language Models}

\author{Amelie Robrecht \\
  Social Cognitive Systems\\
  Bielefeld University \\\And
  Judith Sieker \\
  Computational Linguistics \\
  Bielefeld University \\\And
  Clara Lachenmaier \\
  Computational Linguistics \\
  Bielefeld University \\\AND
  Sina Zarieß \\
  Computational Linguistics \\
  Bielefeld University \\\And
  Stefan Kopp \\
  Social Cognitive Systems \\
  Bielefeld University}

\begin{document}
\maketitle
%\begin{abstract}
%\end{abstract}

\section{Introduction}
 
Within the landscape of linguistic capabilities that have been studied and analyzed in Large Language Models (LLMs), a considerable amount of research has focused on phenomena on the level of morphology and syntax \cite{marvin-linzen-2018-targeted, hu-etal-2020-systematic}.
%Work on probing LLMs often shows that a rich and diverse set of phenomena on the level of morphology and syntax can be learned accurately in LLMs.
Here, the community seems to have agreed on benchmarks and phenomena that an LLM should be capable of (e.g. agreement phenomena \cite{Warstadt2020-kr}). Various studies show that LLMs can handle a rich and diverse set of such phenomena \cite{chang_language_2023}.
Recent inquiries have expanded to investigate the proficiency of LLMs in pragmatic discourse processing \cite{ruis_goldilocks_2022, hu_fine-grained_2023, sieker_beyond_2023}.
%One area of interest has been classification tasks, such as examining whether large language models can comprehend inferences triggered by implicatures or presuppositions, utilizing datasets specifically developed for Natural Language Inference (NLI) tasks \cite{jeretic-etal-2020-natural, kabbara-cheung-2022-investigating}. 

Pragmatic phenomena are often utilized when arguing for or against the reasoning capabilities of LLMs, which are a requirement for grounding in dialog. However, research on pragmatic abilities in LLMs remains more scarce and less systematic. We argue that studying the pragmatic competencies of LLMs is particularly interesting as it bridges aspects of 'core-linguistic' knowledge with the communicative, functional, and contextual aspects of grounding and is still actively discussed in current research \cite{mahowald_dissociating_2024}.  
%What does it mean when LLMs struggle with some pragmatic tasks such as selecting the correct (anti-)presupposition \cite{sieker_when_2023}, but can ... 
% but can compete compete in regards to coherence and outperform humans in terms of informativity \cite{sieker_beyond_2023}. 
What does it mean when models can infer mental states while struggling with implicit meaning \cite{chang_language_2023}? Why do Language Models tussle, especially with phenomena that break language rules, such as humor, irony, and conversational maxims \cite{hu_fine-grained_2023}?

To address these questions and categorize findings effectively, capabilities related to pragmatics and grounding must be mapped out clearly and defined in relation to one another.
%pragmatics must be defined clearly and delineated in scope. %Moreover, within the sphere of pragmatic phenomena, further subdivision into subclasses is essential for a systematic analysis.
In this work, we want to give an overview on which pragmatic abilities have been tested in LLMs so far and how these tests have been carried out. To do this, we first discuss the scope of the field of pragmatics and suggest a subdivision into \textit{discourse pragmatics} and \textit{interactional pragmatics}. We give a non-exhaustive overview of the phenomena of those two subdomains and the methods traditionally used to analyze them. We subsequently consider the resulting heterogeneous set of phenomena and methods as a starting point for our survey of work on discourse pragmatics and interactional pragmatics in the context of LLMs.

\section{Pragmatics in Linguistics}
%%% 
Unlike other linguistic fields, such as syntax or phonetics, which focus on more structured and formal aspects of language, pragmatics encompasses a more heterogeneous set of phenomena that are often less systematic and more context-dependent \cite{ariel_defining_2010}. Negative definitions like the investigation of meaning distinct from pure semantics \cite{cummings2013pragmatics} are fuzzy, and therefore, pragmatics is sometimes even referred to as the garbage can of linguistics \cite{wastebasket}. 
\citet{cummings2013pragmatics} contends that defining pragmatics as the study of how context affects meaning or as language usage analysis is overly broad. Instead, she proposes to define pragmatics as all intentionally expressed meanings that go beyond what is literally said. 
%\citet{cummings2013pragmatics} argues that the obvious attempt to define pragmatics as all the disciplines that study how context affects meaning is inadequate as a definition because it is too broad. According to her, the same applies to the approach that describes pragmatics simply as a subject that analyses how we use language. 
%For \citet{cummings2013pragmatics}, the most suitable definition of the subject area of pragmatics is all intentionally expressed meanings that go beyond what is literally said.
However, numerous endeavours have been made to establish clearer definitions or categorizations within the field.
The Stanford Encyclopedia of Philosophy article on pragmatics, for example, distinguishes between 'classical' and 'contemporary' pragmatics, with classical pragmatics further divided into 'near-side' and 'far-side' \citep{sep-pragmatics}. Near-side pragmatics focuses on explicit content, while far-side pragmatics explores implications beyond literal meanings. Contemporary pragmatics, on the other hand, includes works like Sperber \& Wilson's relevance theory.
%While near-side pragmatics (e.g. David Kaplan's theory on Demonstratives and Stalnaker's account on Context)  examines the facts relevant to determining the explicit content of utterance (including, for example, the resolution of ambiguity and reference), far-side pragmatics (e.g. J. L. Austin and H. P. Grice) focuses on the actions or implications that go beyond the literal meaning, such as speech acts or implicatures.
%The authors classify theories like David Kaplan's on Demonstratives and Stalnaker's on Context as belonging to near-side pragmatics. In contrast, theorists like J. L. Austin and H. P. Grice are argued to be more aligned with far-side pragmatics.
%On the other hand, the "contemporary pragmatics" include, for example, Sperber \& Wilson's work on relevance theory or Levinson's theory of utterance-type-meaning.
Within these categories, \citet{sep-pragmatics} cover several pragmatic phenomena like ambiguity and implicatures. Yet, notably, grounding-relevant phenomena such as turn-taking or repair are overlooked despite being clearly pragmatic in nature. 

Nevertheless, or precisely because of the diversity in the set of pragmatic phenomena, subcategorization is needed.
We propose to cluster them into two main categories: \textit{discourse pragmatics} and \textit{interactional pragmatics}. 
While the \textit{discourse pragmatics} describe formal reasoning processes, including phenomena such as presupposition, implicatures and figurative speech (i.e., aspects of pragmatics that were considered in \citeauthor{sep-pragmatics}'s article and could be described as near-side pragmatics), the \textit{interactional pragmatics} address conversational reasoning phenomena, such as politeness, turn taking or repair (which could be designated as far-side pragmatics).
%%%%%%% Discourse pragmatics 
Discourse pragmatics is often addressed in classical pragmatics and Natural Language Processing. The phenomena are mostly connected to text coherence. They can be found in a dialog but do not require direct interaction. These phenomena have been in the center of attention for decades. Often, testing instruments -- drawing from the field of psychology or psycholinguistics \cite{ettinger2020-bert, sieker_when_2023} -- are established. Additionally, theories from discourse pragmatics provide frameworks to describe these phenomena \cite{frank_predicting_2012, degen_rational_2023}.

%%%%%%% Interactional pragmatics 
Besides, there is a field of pragmatics that we refer to as interactional pragmatics. Here, the focus is rather on the interlocutors' interplay. A lot of research has been done on conversation analysis \cite{sacks1978simplest, atkinson1984structures} or politeness theory \cite{brown1987politeness, goffman1955face, leech2014pragmatics}. 
Conversation analysis utilises a strictly qualitative methodology borrowed from sociology and addresses the issue of "how we use language" at its core. The investigations on natural data focus on the organising principles that underlie human communication \cite{sacks1978simplest, atkinson1984structures}.
In politeness theory, nuances of spoken language are emphasized\cite{brown1987politeness}.
Research in computer science and computational linguistics often addresses similar questions from the perspective of human-robot interaction (HRI). \citet{kumar_politeness_2022} reveal the positive impact politeness has on the enjoyment, satisfaction and trust participants perceive in an interaction with a robot. And \citet{skantze_turn-taking_2021} give an overview of research on turn-taking behavior in HRI. Also, further interactional phenomena such as adaptation \cite{robrecht_study_2023, axelsson_you_2023, stange_tell_2022} or grounding \cite{jung_affective_2017} have been subject to manifold approaches and studies in the field.

\section{Approaches to pragmatics in LLMs}

%%%%%% Discourse pragmatics in LLMS
There are various examples of research that tests discourse pragmatic reasoning capabilities in language models.
\citet{ruis_goldilocks_2022} investigate the extent to which LLMs such as OPT, T5 or GPT-4 may understand conversational implicatures.
In-scale and between-scale scalar inferences in BERT are tested by comparing the model's abilities to the human performance by \citet{hu_expectations_2023}. \citet{carenini_large_2023} take a look at the understanding GPT-2 has of metaphors, explaining their results using the Rational Speech Act theory. \citet{hu_fine-grained_2023} test seven discourse pragmatic phenomena (including maxims, metaphor, and coherence) in different versions of GPT-2, GPT-3 and T5.
%ToDo: List of phenomena
%%%%%%% Interactional pragmatics in LLMs
Moreover, the outcomes appear less promising when examining the study of interactional pragmatics in LLMs, the pragmatic category which covers most of the grounding-related phenomena. As this field of pragmatics is not as settled and the phenomena are harder to analyze due to their close connection to interaction, spoken language and spontaneous adaptation, there is a lack of instruments and measurements. \citet{milicka_large_2024} show that GPT-3 and GPT-4 are able to decrease their cognitive abilities to simulate other personas. Also \citet{wilf_think_2023} test the perspective-taking abilities of GPT-3, GPT-4, and Llama2, using chain-of-thought prompting. 
Nevertheless, most research connected to interactional pragmatics focuses on Theory of Mind or related theories \cite{gandhi_understanding_2023, wilf_think_2023}. It remains questionable whether these phenomena should be considered part of interactional pragmatics or not. 
%TODO: Paper zu repair, politeness, turn-taking in LLMs

%Conclusion/ Was wir planen
\section{Contribution}
We argue that there is a need for a more precise definition of pragmatic capabilities in research that studies the communicative behavior of LLMs. As a first step, we propose to distinguish discourse and interactional pragmatic abilities, for which we will discuss classification criteria and borderline cases. Further, we summarize which pragmatic phenomena have been tested in LLMs, how they are related to grounding, which methodology has been used, and which models have been considered.
\section{Acknowledgments}
Amelie Robrecht's and Stefan Kopp's research was funded by the Deutsche Forschungsgemeinschaft (DFG, German Research Foundation): TRR 318/1 2021 – 438445824.
Judith Sieker, Clara Lachenmaier and Sina Zarrieß received financial support from the project SAIL: SustAInable Life-cycle of Intelligent Socio-Technical Systems: NW21-059A, funded by the program “Netzwerke 2021” of the Ministry of Culture and Science of the State of North Rhine-Westphalia, Germany.

\bibliography{references, custom2}

%\bibliography{latex/custom}s

\end{document}